\documentclass[journal]{IEEEtran}
\IEEEoverridecommandlockouts
\usepackage{cite}
\usepackage[T1]{fontenc}
\usepackage{amsmath,amssymb,amsfonts}
\usepackage{graphicx}
\usepackage{algorithmic}
\usepackage{textcomp}
\usepackage{hyperref}
\usepackage{url}

\usepackage{CJKutf8}
\usepackage{subfigure}
\usepackage{tabularx}
\usepackage{booktabs}
\usepackage{multirow}
\usepackage{xcolor}
\usepackage{wrapfig}

\usepackage{amsthm,amsmath,amssymb}
\usepackage{mathrsfs}
\def\BibTeX{{\rm B\kern-.05em{\sc i\kern-.025em b}\kern-.08em
    T\kern-.1667em\lower.7ex\hbox{E}\kern-.125emX}}

\begin{document}

\title{ EmMixformer: Mix transformer for eye movement recognition\\

}

\author{
       Huafeng Qin,
         Hongyu Zhu,
         Xin Jin,
         Qun Song,
         ~\IEEEmembership{Senior member, IEEE} Mounim~A.~El-Yacoubi, and ~\IEEEmembership{Fellow, IEEE} Xinbo Gao .
          
\thanks{H. Qin, H.Zhu, X. Jin, and Q. Song is with Chongqing intelligence perception and block chain technology  key laboratory,  the School of Computer Science and Information Engineering, Chongqing Technology and Business University, Chongqing 400067, China (e-mail: qinhuafengfeng@163.com).} 
\thanks{M. A. El-Yacoubi is with SAMOVAR, Telecom SudParis, Institute Polytechnique de Paris, 91120 Palaiseau, France (e-mail: mounim.el\_yacoubi@telecom-sudparis.eu).}
\thanks{ X. Gao is with the Chongqing Key Laboratory of Image Cognition, Chongqing University of Posts and Telecommunications, Chongqing 400065, China (e-mail: gaoxb@cqupt.edu.cn).}

\thanks{Manuscript received June XX, 2023; revised XXXX XX, 201X. This work was supported in part by the National Natural Science Foundation of China (Grant Nos. 61976030, 62072061, and U20A20176), the Scientific Innovation 2030 Major
Project for New Generation of AI (Grant No. 2020AAA0107300), the Fellowship of China Post-Doctoral Science Foundation (Grant No. 59676651E), and the
Science Fund for Creative Research Groups of Chongqing Universities (Grant No. CXQT21034, Grant Nos. KJQN201900848 and KJQN201500814). }}
\maketitle

\begin{abstract}

Eye movement (EM) is a new highly secure biometric behavioral modality that has received increasing attention in recent years. Although deep neural networks, such as convolutional neural network (CNN), have recently achieved promising performance, current solutions fail to capture local and global temporal dependencies within eye movement data. To overcome this problem, we propose in this paper a mixed transformer termed EmMixformer to extract time and frequency domain information for eye movement recognition. To this end, we propose a mixed block consisting of three modules, transformer, attention Long short-term memory (attention LSTM), and Fourier transformer. We are the first to attempt leveraging transformer to learn long temporal dependencies within eye movement. Second, we incorporate the attention mechanism into LSTM to propose attention LSTM with the aim to learn short temporal dependencies. Third, we perform self attention in the frequency domain to learn global features. As the three modules provide complementary feature representations in terms of local and global dependencies, the proposed EmMixformer is capable of improving recognition accuracy. The experimental results on our eye movement dataset and two public eye movement datasets show that the proposed EmMixformer outperforms the state of the art by achieving the lowest verification error. 

\end{abstract}


\section{Introduction}
Biometrics refers to identifying or verifying person’s identity
based on their biometric traits\cite{Jain2008AnIT}. Unlike traditional non-biometric methods such as key, password, and card, biometric traits are difficult to copy and forge, and cannot be forgotten, making them highly secure and convenient. Current biometric authentication techniques can be categorized into two types: Physiological and Behavioral modalities \cite{Qin2023LabelEM}. The former utilize static body biometrics, such as fingerprint recognition \cite{Jirachaweng2011ResidualOM}, face recognition \cite{Wang2010FaceRU}, vein recognition \cite{Qin2011RegionGF}, and iris recognition\cite{Roy2011TowardsNI}, while the latter are related to user dynamic behavior, such as eye movements\cite{Lohr2021EyeKY}, gait \cite{Cola2016GaitbasedAU},  signature\cite{Kiran2021OfflineSR}, and voice\cite{Liang2020BehavioralBF}. Physiological modalities are intrusive and may require user's cooperation, which may be inconvenient. Additionally, they are prone to being recreated or captured, in addition to being vulnerable to spoofing as they can be artificially synthesized by using, for instance, gummy fingers\cite{Chugh2018FingerprintSB} or deepfake technology. Dynamic biometric traits, by contrast, exploring active and dynamic personal characteristics, are unique and almost impossible to spoof since they are caused by subconscious processes. Recently, behavioral biometrics, such as eye movement with temporal information associated with human cognition, have received increasing focus in biometrics. 

 \subsection{The principle of Eye movement}
 Eye movement biometrics \cite{Lohr2021EyeKY} aims to identify the behavioral patterns and information regarding physiological properties of tissues and muscles generating eye movement. These behavioral and physiological patterns provide rich information about the cognitive brain functions and neural signals controlling eye movements.  Compared to other biometrics, eye movement biometrics technology has significant advantages. It not only provides liveness detection \cite{Makowski2020BiometricIA}, ensuring thereby security even in unconscious states, but its reliance on both the properties of the oculomotor plant and the neurological properties controlled by the brainstem makes eye movements almost impossible to replicate  \cite{Holland2013ComplexEM}. Eye movement biometrics, in addition, are highly suitable for multi-modality recognition as it may be seamlessly combined with other eye-related physiological traits such as iris and pupil\cite{Blignaut2013MappingTP}. These key advantages explain why eye movement biometrics has become a hot research topic over the recent years.



\begin{figure}[htbp]
    \centerline{\includegraphics[scale=0.5]{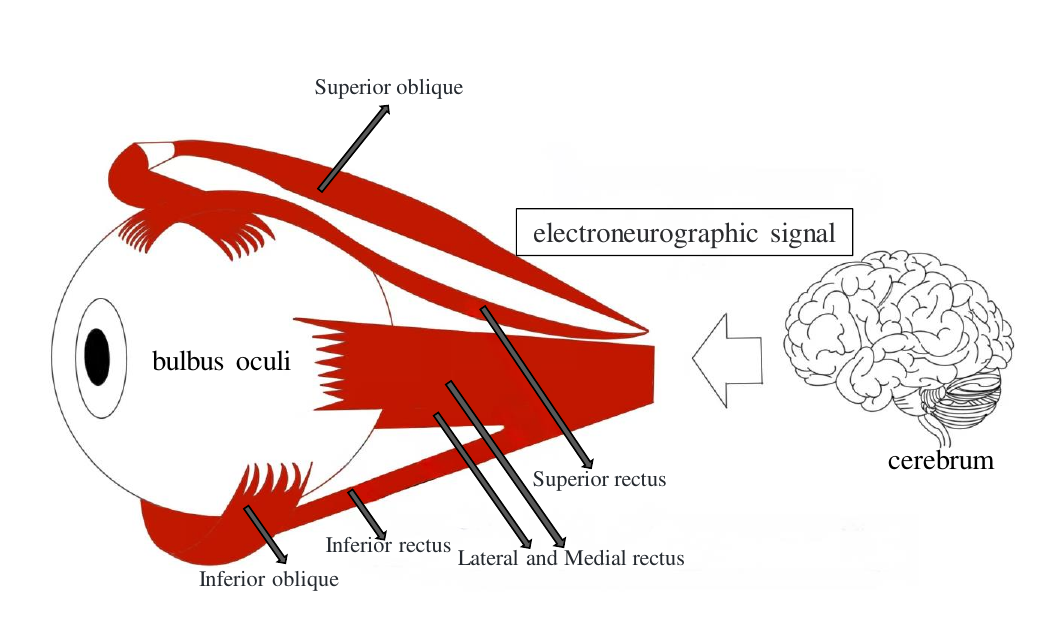}}
    \caption{Eye Musculature.}
    \label{fig1}
\end{figure}

The human eye moves within six degrees of freedom. Oculomotor movements encompass the coordinated contractions and relaxations of the six eye muscles (four rectus and two obliques), controlled by the brainstem nerves during visual processing (Fig.\ref{fig1}). These movements serve as a vital mechanism for humans to selectively extract information from the visual environment. As they enable us to focus our attention on objects of interest, eye movements, like keystroke dynamics\cite{Gunetti2005KeystrokeAO}, gait \cite{Cola2016GaitbasedAU} and electrocardiography (ECG)\cite{Zheng2011AnEU}, contain rich information about cognitive functions, neural pathways, and other brain-related aspects \cite{George2016ASL}. The literature shows that the way we move our eyes in response to a given stimulus is highly individual \cite{noton1971scanpaths,Nagamatsu2008OnepointCG} and that these individual characteristics are reliable over time \cite{Akkil2014TraQuMeAT}. Hence, eye movements have been proposed as a behavioral biometrics \cite{Kasprowski2004EyeMI,bednarik2005eye}.

Eye movements can be divided into saccades and fixations \cite{Holland2012BiometricVV}. Saccades involve rapid eye movements as the gaze transitions from one point to another, typically occurring at speeds of 200-400°/s, and peak velocities that may exceed 600°/s\cite{Rigas2018StudyOA}. To initiate a saccade, related neural circuitry calculates the disparity between the starting and target positions. It then sends precise neural signals to the extra-ocular muscles, which in turn rotate the eye accordingly. If the eyeball fails to accurately reach the intended target, the muscles will make minute corrections. Fixations, by contrast, relatively  focus on a specific gaze point while exhibiting high-frequency, low-amplitude nystagmus movements, and low-velocity eye movements with a typical duration of 100-400 ms. Eye tracking devices can capture both position and movement data for saccades and fixations, which can be leveraged to extract eye movement features for biometric identification.


\subsection{Related Work}

\paragraph{ Traditional eye movement recognition approaches}
Traditional approaches use handcrafted descriptors and machine learning (ML) algorithms for feature extraction and recognition. Kasprowski and Ober\cite{Kasprowski2004EyeMI} made a first attempt to explore eye movement features for identification. Their experimental results on a small dataset showed that eye movement comprises discriminating information, allowing the classification model to achieve low verification error. In 2011, Holland and Komogortsev \cite{Holland2011BiometricIV} gathered eye-movement scan path data generated by 32 volunteers and designed 15 eye movement metrics for feature extraction. A weighted average method was then used to improve accuracy. In 2012, Komogortsev et al. \cite{Komogortsev2012BiometricAV} introduced the concept of oculomotor plant characteristics (OPC) for identification purposes and employed Student’s t-test enhanced by voting and Hotelling's T-Square test to distinguish between gaze and sweep events. Subsequently, they \cite{Holland2013ComplexEM} extracted  more intricate features such as eye hopping abnormalities, compound eye hopping, and dynamic overshooting from eye movements, input to Random Forest (RF) and Support Vector Machines (SVM) for classification. In 2017, Bayat and Pomplun \cite{Bayat2017BiometricIT} extracted a feature set comprising 28 features for identification by using gaze, sweep, and pupil size information. They combined three classifiers and evaluated performance on a dataset of 40 objects. Similarly, Cuong et al. \cite{Nguyen2012MelfrequencyCC} proposed a method to encode eye movement using Mel-Frequency Cepstrum Coefficients (MFCCs) without distinguishing between sweeping and gazing. In 2018, Li et al. \cite{li2018biometric} proposed a multi-channel Gabor wavelet transform to extract texture features from eye movement trajectories and SVM was employed for recognition. In \cite{lohr2020metric}, Lohr et al. proposed three multilayer perceptrons (MLP) with triplet loss to learn the features of three event types, fixation, saccade, and post-saccadic oscillations (PSO). In recent years, Long Short-Term Memory (LSTM) was proposed to extract temporal features for gaze pattern recognition\cite{yeamkuan2020fixational,Jia2018BiometricRT}.

\paragraph{ Deep Learning based eye movement recognition approaches}
Recently, deep learning (DL) models, e.g. Convolutional Neural Networks (CNNs), thanks to their robust feature representation, were successfully applied for image segmentation\cite{Shelhamer2014FullyCN}, image classification\cite{Krizhevsky2012ImageNetCW}, object detection\cite{Girshick2013RichFH}, and, recently, eye movement recognition. Lohr et al. \cite{Lohr2021EyeKY}, for instance, introduced an exponentially-dilated CNN with multi-similarity loss for user authentication. Taha et al. \cite{Taha2023EyeDriveAD} combined LSTM and dense networks to learn temporal characteristics from the eye movement (EM) profiles and proposed an end-to-end learning model for driver authentication. Likewise, \cite{Jger2019DeepEB} developed a CNN to learn feature representation from the raw eye-tracking signal. For the same purpose, Makowski et al. \cite{Makowski2021DeepEyedentificationLiveOB} investigated a CNN to process binocular eye tracking signals and the relative positions of the stimuli to detect replay attacks. In 2022, Makowski et al. \cite{Makowski2022OculomotoricBI} collected an eye tracking dataset from 66 participants in sober, fatigued and alcohol-intoxicated states, and fine-tuned the DeepEyedentification \cite{Jger2019DeepEB} model to investigate the effects of drunkenness and fatigue on eye movement biometrics.  At the same year, D. Lohr et al. \cite{Lohr2022EyeKY} proposed a DenseNet network architecture for end-to-end eye movement biometric authentication.

\subsection{Motivation and Contribution}
 Traditional approaches use manually-designed algorithms to extract, from eye movement, handcrafted features, input to ML approaches such as SVM and MLP, which suffers from the following issues: 1) Manually-designed models rely on human prior assumptions such as Gaussian distributions. These assumptions, however, are not always effective to learn robust eye movement patterns as the latter's underlying data distributions are usually much more complex in real-life applications. This explains also why it is hard to design a mathematical model to effectively model such distributions; 2) Traditional ML approaches such as SVM and RF can be viewed as a neural network with very few (one or two) layers for feature representation learning. Their representation capacity is therefore relatively limited.  DL models trained on huge data, by contrast, are capable of learning high-level features by stacking a number of layers. For eye movement recognition, DL models, such as CNN, have recently achieved promising performance compared to ML approaches and become an increasing modeling trend.  CNN  can effectively capture local patterns and features, such as edges, corners, textures, or shapes, by sharing weights and local perception. CNN, however, is difficult to capture long temporal dependencies within sequences because of limited receptive field\cite{Guo_2022_CVPR}\cite{Vaswani2017AttentionIA}. As eye movement is typically temporal data, collected over time with observations recorded at different time points, existing DL-based approaches is difficult to extract effective features that embed such long-range time dependencies. Recently, transformers with attention mechanisms have emerged as new start-of-the art classifiers thanks to their high capacity to capture long dependencies among tokens, with impressive performance obtained, first, in various natural language processing (NLP) tasks \cite{Vaswani2017AttentionIA} and then in computer vision \cite{Dosovitskiy2020AnII}. In fact, the capacity of model to learn long dependencies is related to the length of the paths forward and backward signals. The shorter paths between any combination of positions in the input and output sequences allow to learn longer range dependencies\cite{hochreiter2001gradient}. For example, the  work  \cite{Vaswani2017AttentionIA} compares the maximum path length between any two input and output positions in networks such as transformer, LSTM, and CNN  and shows that transformer has the shortest paths and is capable of learning long range dependency.

Inspired by these facts, we propose in this paper a novel mix transformer for eye movement recognition, named EmMixformer. EmMixformer includes three modules, transformer, attention-LSTM, and Fourier-transformer. The eye movement signal is first split into fast and slow velocity signals, separately input to encoder for feature embedding. The embedding features are then input to the mix transformer module to learn a robust representation for classification. The transformer and attention LSTM aims at learning 
temporal dependencies within the eye movement signal, while the Fourier transformer learns a global feature representation. Our mix transformer can, therefore, learn complementary features. The contributions of our work are summarized as follows:
\begin{itemize}
    \item We are the first to propose transformer for the eye movement recognition task by proposing an eye movement mix transformer (EmMixformer) to learn a robust feature representation.  Concretely,  we investigate a mix block, consisting of three modules (as shown in Fig.\ref{fig2}): a transformer, an attention LSTM and a Fourier-transformer. Our approach is capable of extracting eye movement features in both time and frequency domains.
    
    \item We design an attention LSTM to learn robust temporal dependencies by incorporating the attention mechanism into LSTM structure. Our resulting  attention-based LSTM model is capable of capturing robust features embedding temporal dependencies for eye movement recognition.
    
    \item We propose a Fourier transformer to learn global dependencies by performing self attention in the frequency domain. This is the first transformer to perform frequency domain learning. 
    \item We build a low-frequency eye tracking dataset acquired thanks to a state-of-the-art portable eye-tracking instrument (Tobii Pro glasses3) and make it publicly available to promote research on eye movement recognition. The data captured by this low cost sensor are more realistic, which promotes more valuable benchmarking in the area of eye movement recognition.
    \item We have carried out rigorous experiments on our eye movement dataset and on two public eye movement datasets to assess our approach. The experimental results show that our approach outperforms existing works in terms of reducing the verification error, and achieves new state-of-the-art-recognition results. 
\end{itemize}

\section{Methodology}
\begin{figure}[htbp]
    \centerline{\includegraphics[scale=0.82]{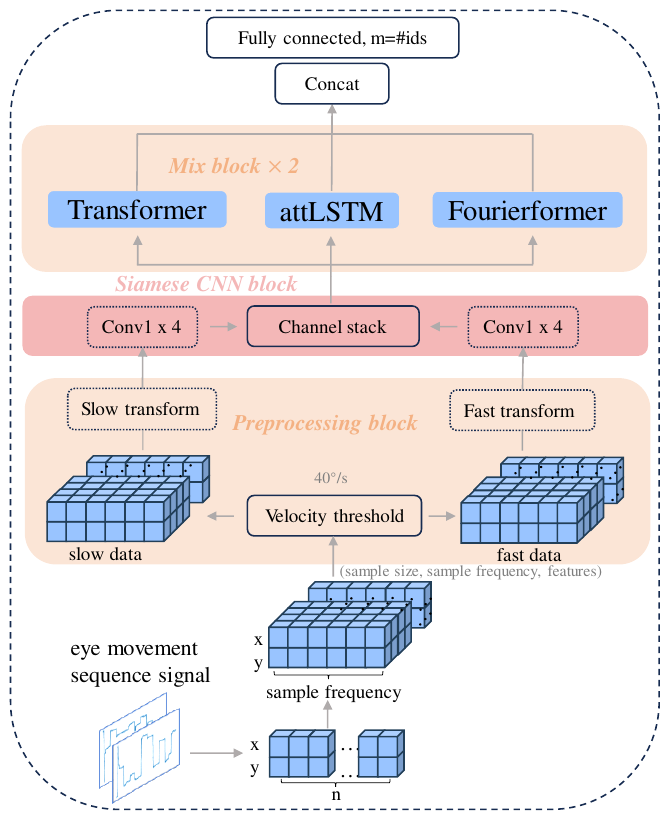}}
    \caption{The framework of the proposed EmMixformer model}
    \label{fig2}
\end{figure}

In this section, we detail the  proposed EmMixformer model, as illustrated in Fig.\ref{fig2}. EmMixformer comprises a prepossessing block, a Siamese CNN block, and a Mix block. The prepossessing blorck splits the input eye movement data into low and fast velocity data. The Siamese CNN block comprises two sub-nets for feature embedding of the two velocity data, while the Mix block comprises three modules, transformer, attention LSTM (attLSTM) and Fourier transformer (Fourierformer). The slow and fast velocity signals output by the prepossessing block are input to the Siamese CNN block for feature embedding. The resulting features are forwarded to the mix block for feature extraction and recognition. We detail, in this section, the Siamese CNN block and the Mix block, and provide the details of the prepossessing block in Section III.   

\subsection{Siamese CNN block}
The Siamese CNN block consists of two sub-networks, each stacking four one-dimensional convolutional modules for feature extraction. {The size of the convolution kernel increases to obtain a progressively larger receptive field.} Each module comprises a convolutional layer,  a batch normalization layer, a ReLU layer, and an average pooling layer. The two subnets take respectively the preprocessed fast and slow eye movement signals as inputs and output the corresponding embedding feature representations. We then concatenate the output feature representations from the two subnets to obtain the complete feature representation. {The $x$ and $y$ represent the horizontal and vertical  coordinate sequences of eye movement data, and $n$ represents the sequence length.}

\subsection{Mix block}
The Mix block comprises three modules, an attention LSTM (attLSTM), a Transformer, and a Fourier-transformer (Fourierformer). The three modules take the Siamese CNN block's output as input to extract rich and  comprehensive temporal-spatial features. Concretely, a temporal sequence from the Siamese CNN block is divided into a number of sub-sequences, which are input to attLSTM and transformer to extract the temporal features.  The attLSTM module learns short-term dependencies within the temporal sequences, while the transformer module extracts long-term dependencies within the temporal sequences thanks to its self-attention mechanism \cite{Liu2021SwinTH}. Their combination, therefore, can learn a comprehensive temporal feature representation. The third module, Fourierformer, takes as input the temporal sequences to learn a global frequency-domain feature representation. As a result, our mix block extracts both a feature representation embedding temporal dependencies and a global feature representation, thus effectively improving the modeling of time-series eye movement classification.

\paragraph{Transformer}
Transformer, introduced in 2017 by Vaswani et al \cite{Vaswani2017AttentionIA}, has become the state of the art for NLP \cite{Vaswani2017AttentionIA} and more recently computer vision \cite{Liu2021SwinTH}. Unlike traditional sequence models such as recurrent neural networks (RNNs) or CNNs, Transformer relies solely on a key idea, the self-attention mechanism, eliminating the need for sequential processing. The self-attention mechanism allows the model to weigh the importance of different tokens in a sequence, and enables it to capture effectively long-range dependencies and context information. By attending to all tokens simultaneously, Transformer can process sentences in parallel, leading to significantly faster training and inference processing times, compared to existing sequential models \cite{Liu2021SwinTH}.

For eye movement sequence $X$=($x_{1}$,..., $x_{T}$), position encoding is introduced by adding the position information into the embedding vector. We treat each element in the eye movement sequence as a token, which yields $T$ tokens for $X$. To learn the temporal feature representation, we perform self-attention among tokens.  We transform the $t$th token $x_{t}$ into three vectors, query $q_t$, key $k_t$, and value $v_t$, by three weight matrices $w_q$, $w_k$, and $w_v$. We pack the $T$ query vectors  $q_1,q_2,...,q_T$ of the input sequence into a matrix $Q$. Similarly, the keys and values are packed together to obtain matrix $K$ and $V$.  The self-attention among $T$ tokens is implemented by Eq.(\ref{eq1}).
 
\begin{equation}
    Attention\left ( Q,K,V\right )=Softmax\left ( \frac{QK^{T}}{\sqrt{d_{k}}}\right )V
    \label{eq1}
\end{equation}
where Softmax($\cdot$) is applied to the rows of similarity matrix $QK^{T}$, and $d_{k}$ provides normalization. Query $Q$, keys $K$, and values $ V$ are the projections of the tokens computed by projection matrices $W_q $, $W_k $, and $W_v$, respectively.

To learn a robust feature representation, we set a multi-head self attention by concatenating $L$ attention heads (Eq.(\ref{eq2})).
\begin{equation}
    MultiHead=Concat\left ( head_{1},...,head_{L}\right ) W
        \label{eq2}
\end{equation}
where $L=4$, $head_{l}=Softmax\left ( \frac{Q_lK_l^{T}}{\sqrt{d'_{k}}}\right )V_l$ denotes the output of the $l$th attention head, and $W$ is the linear transformation matrix. Subsequently, the representation $ MultiHead$ is subjected to residual concatenation and layer normalization operations by  Eq.(\ref{eq3}):
\begin{equation}
   Z=\left ( X+ MultiHead(LN(X))\right )
    \label{eq3}
\end{equation}

Finally the representation of each node is non-linearly transformed by a feed-forward neural network with two layers:
\begin{equation}
     X_{transformer}=  LN(MLP(LN(Z))+Z)
     \label{eq4}
\end{equation}

To simplify representation, we combine Eq.(\ref{eq1}), Eq.(\ref{eq2}), Eq.(\ref{eq3}), Eq.( \ref{eq4}) to obtain Eq.( \ref{eq5}).  
\begin{equation}
     X_{transformer}=  transformer(X)
     \label{eq5}
\end{equation}

\paragraph{Attention LSTM}
LSTM\cite{Hochreiter1997LongSM} is a special kind of RNN that can learn long-term dependencies in a more efficient way than RNN, by introducing memory cells and multiple control gates. LSTM is widely used for sequence processing \cite{Ordonez2016DeepCA} \cite{Ngo2017SaccadeGP}. The key component of an LSTM cell is the memory cell, responsible for storing and updating information over time. A standard LSTM cell consists of three control gates: forget gate, input gate, and output gate. These gates control the flow of information by using a sigmoid function as an activation function. The input gate determines how much of the new input should be stored in the memory cell, while the forget gate decides how much of existing information should be discarded. The output gate regulates the amount of information that is passed to the next time step or output. The LSTM architecture allows gradients to flow through the network for longer periods of time, enabling it to capture long-term dependencies in the data\cite{Greff2015LSTMAS}. In order to make better use of cell state information and improve the expression and memory ability of the model, Peephole connection \cite{Gers2000LearningTF} has been introduced into the LSTM unit. Specifically, cell state$ c_t$ and hidden state $ h_t$ from the previous step are used as additional parameters for the update forget gate $f_{t}$ and input gate $i_{t}$: the four gates are calculated by the following formulas:
\begin{equation}
f_{t}=\sigma \left ( W_{f}{x}+U_{f}h_{t-1}+V_fC_{t-1}+b_{f}\right )
\label{eq6}
\end{equation}

\begin{equation}
i_{t}=\sigma \left ( W_{i}{x}+U_{i}h_{t-1}+V_iC_{t-1}+b_{i}\right )
\label{eq7}
\end{equation}

\begin{equation}
o_{t}=\sigma \left ( W_{o}{x}+U_{o}h_{t-1}+V_oC_t+b_{o}\right )
\label{eq8}
\end{equation}

\begin{equation}
\widetilde{C_{t}}=tanh \left ( W_{C}{x}+U_{C}h_{t-1}+b_{C}\right )
\label{eq9}
\end{equation}

\begin{equation}
C_{t}=f_{t}\odot C_{t-1}+i_{t}\odot\widetilde{C_{t}}
\label{eq10}
\end{equation}

\begin{equation}
h_{t}=o_{t} * tanh({C_{t}})
\label{eq100}
\end{equation}
where $f_{t}$ , $i_{t}$ and $o_{t}$ are the forgetting gate, input gate and output gate, respectively. $\widetilde{C_{t}}$ , $C_{t}$ and $h_{t-1}$ are the candidate cell state, the cell state, and the hidden state at the last moment, respectively.
 $\odot$ denotes the element-by-element matrix multiplication {and $\sigma$ represents the sigmoid activation function, respectively.}
\begin{figure}[htbp]
    \centerline{\includegraphics[scale=0.6]{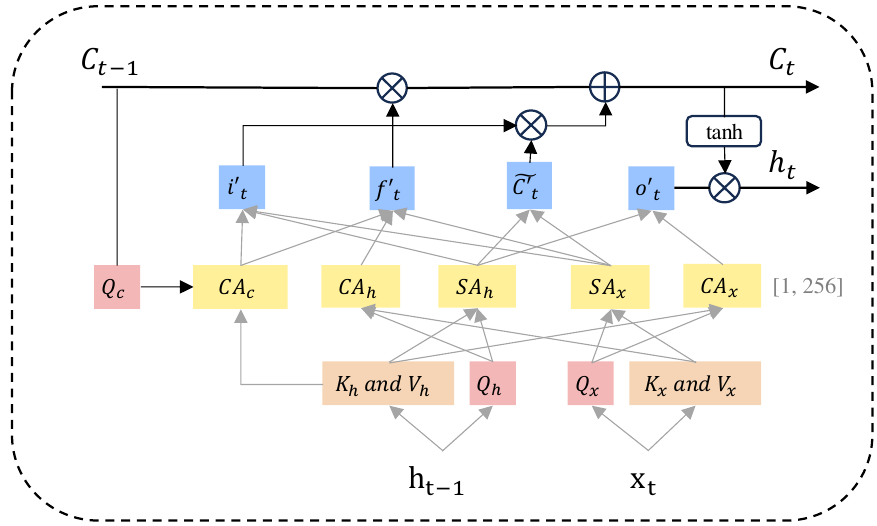}}
    \caption{ Attention LSTM}
    \label{fig2b}
\end{figure}

To improve LSTM learning capacity, we incorporate the attention mechanism to propose attention LSTM, as shown in Fig. 3. Let $x$, $h'_{t-1}$ and $C'_{t-1}$ be the input vector, previous hidden state and previous cell state in our attention LSTM. We calculate query matrix $Q_x$, key matrix $K_x$, and value matrix $V_x$ of $x_t$. Similarly, we obtain query matrix $Q_h$, key matrix $K_h$, and value matrix $V_h$ of $ h_{t-1}$, and  query matrix $Q_c$, key matrix $K_c$, and value matrix $V_c$ of $C_{t-1}$. Based on Eq(\ref{eq1}), the self attention of input vector $x_t$ is computed by Eq. (\ref{eq11}): 

\begin{equation}
\begin{gathered}
SA_{x}=Attention(Q_{x},K_{x},V_{x})\\
 \end{gathered}
 \label{eq11}
\end{equation}

Similarly, the self attention of hidden state $h_{t-1}$, and the cross attention between $x_t$ and $h_{t-1}$ are calculated by:

\begin{equation}
\begin{gathered}
SA_h=Attention(Q_h,K_h,V_h),
 \end{gathered}
 \label{eq12}
\end{equation}

\begin{equation}
\begin{gathered}
CA_{x}=Attention(Q_{x},K_h,V_h),
 \end{gathered}
 \label{eq13}
\end{equation}

\begin{equation}
\begin{gathered}
CA_h=Attention(Q_h,K_{x},V_{x}),
 \end{gathered}
 \label{eq14}
\end{equation}

\begin{equation}
CA_c=Attention(Q_c,K_h,V_h),
 \label{eq15}
\end{equation}
where $CA_{x}$ and $CA_h$ denote cross attention between $x_t$ and $h_{t-1}$, $CA_c$ is the cross-attention vector of cell state $C_{t-1}$ and hidden state $h_{t-1}$, while $SA_h$ and $SA_x$ represents the self attention of the hidden state and input. 

The four gates in our attention LSTM are computed by:
\begin{equation}
i'_{t}=\sigma\left ( linear(Concat\left ( SA_{x},SA_{h},CA_c\right ))\right )
\label{eq17}
\end{equation}

\begin{equation}
f'_{t}=\sigma\left ( linear(Concat\left ( CA_{h},SA_{x},CA_c\right ))\right )
\label{eq19}
\end{equation}

\begin{equation}
o'_{t}=\sigma\left ( linear(Concat\left ( CA_{x},SA_{h}\right ))\right )
\label{eq20}
\end{equation}

\begin{equation}
\widetilde{C'_{t}}=tanh \left ( linear(Concat\left ( SA_{x},SA_{h}\right ))\right )
\label{eq18}
\end{equation}

\begin{equation}
C'_{t}=f'_{t}\odot C'_{t-1}+i_{t}\odot\widetilde{C'_{t}}
\label{eq21}
\end{equation}

\begin{equation}
h'_{t}=o'_{t} * tanh({C'_{t}})
\label{eq101}
\end{equation}

The LSTM  takes sequence ($x_{1}$,..., $x_{t}$) with  256 dimensions and outputs hidden state sequence ($h'_{1}$,..., $h'_{T}$). 
To simplify the description, we pack the Eq.(\ref{eq11}-\ref{eq21}) into Eq.(\ref{eq22}):
  
\begin{equation}
X_{attLSTM}=attLSTM\left ( x_1,x_2,..., x_t\right )
\label{eq22}
\end{equation}

\begin{figure}[htbp]
    \centerline{\includegraphics[scale=0.8]{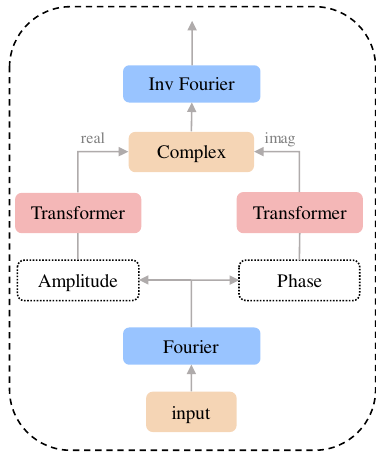}}
    \caption{ Fourier transformer}
    \label{fig4}
\end{figure}

\paragraph{Fourier transformer}
To learn a global feature representation, we perform self attention in the frequency domain by Fourier transformation (Fig.\ref{fig4}). Based on Fourier theory \cite{Li2023LocalGlobalTE} \cite{Zhou2022DeepFU}, feature learning in frequency domain has an image-wide receptive field by channel-wise Fourier transformation. 

A one-dimensional discrete Fourier transform (DFT) transforms sequence $X$ from time domain to frequency domain (Eq.(\ref{eq23})):
\begin{equation} 
\mathcal{F}(X)\left (k \right )=\frac{1}{\sqrt{T}}\sum_{t=0}^{T-1}X(t)e^{-j\frac{2\pi}{T}kt}
 \label{eq23}
\end{equation}
where $\mathcal{F}(X)\left (k \right )$ is the complex value of the $k$-th frequency component in the frequency domain, $X(t)$ is the complex value of the $t$th sampling point in the time domain, $T$ is the sequence length, and $\frac{1}{\sqrt{T}}$ ensures normalized consistency between the DFT and the inverse Fourier transform (IDFT).

Using the real part $Re(X)$ and the imaginary part $Im(X)$ of $\mathcal{F}(X)\left (k \right )$, we obtain the amplitude component $A(X) $ and the phase component $\varphi \left ( X\right )$  by the following equations:

\begin{equation}
A(X)\left ( v\right )=\sqrt{Re\left ( X \right ) \left ( k\right )^{2}+Im\left ( X\right )\left (k\right )^{2}}
\label{eq24}
\end{equation}

\begin{equation}
\varphi \left ( X\right )\left ( k\right )=arctan\left ( \frac{Im\left ( X\right )\left ( k\right )}{Re\left ( X\right )\left ( k\right )}\right )
\label{eq25}
\end{equation}

Next, we extract the feature representation from amplitude component $A(X)\left ( k\right )$ and phase component $\varphi(X) \left ( k\right )$ using two independent transformers. The self attention operation is performed to capture the  amplitude information  $ A_{transformer}$ and phase information $\varphi_{transformer}$, computed as:

\begin{equation}
     A_{transformer}=  transformer(A(X))
    \label{eq26}
\end{equation}

\begin{equation}
     \varphi_{transformer}=  transformer(\varphi(X))
    \label{eq27}
\end{equation}

Subsequently, the feature representations of the amplitude and phase components are combined into complex numbers and transformed back into the spatial domain by IFFT 

\begin{equation}
X'\left ( t\right )=\mathcal{F}^{-1}\left (complex(  A_{transformer},\varphi_{transformer}) \right )
\label{eq28}
\end{equation}

Finally, we concatenate the outputs of the three modules, i.e. attention LSTM, Transformer, and Fourier transformer, to obtain the mix block (Eq.(\ref{eq29})), with the mixing block stacked in two layers.

\begin{equation}
 Y= Concat(X_{attLSTM},X_{transformer}, X' )
 \label{eq29}
\end{equation}

\section{Experimental results}

To assess our approach, we have carried out extensive experiments on our self-built dataset and two public datasets, GazeBase \cite{Griffith2020GazeBaseAL} and JuDo1000 \cite{Makowski2021DeepEyedentificationLiveOB}. We compare our approach with state-of-the-art works, namely DEL\cite{Makowski2021DeepEyedentificationLiveOB}, Expansion CNN\cite{Lohr2021EyeKY}, Dense LSTM\cite{Taha2023EyeDriveAD} and DenseNet \cite{Lohr2022EyeKY}. All experiments were conducted in Pytorch and implemented on a high performance computer with a NVIDIA GTX 3090Ti GPU.

\subsection{Dataset}
\begin{table}[!htbp]
\caption{ Various eye movement datasets}
\centering
\begin{tabular}{c c c c c} 
\toprule
 \textbf{Data set} &\textbf{Device} & \textbf{Freq.(Hz)} & \textbf{Chin rest} & \textbf{Subj.} \\ [0.8ex] 
\midrule
 EMglasses & Tobii pro glasses3 & 50 & no  & 203 \\ 
 JuDo1000\cite{Makowski2021DeepEyedentificationLiveOB} & EyeLink Portable & 1000 & yes & 150 \\
 GazeBase\cite{Griffith2020GazeBaseAL} & EyeLink 1000 & 1000 & yes  & 322 \\
 
 CEM-\begin{CJK*}{UTF8}{gbsn}Ⅰ\end{CJK*}\cite{Holland2012BiometricVV} & Tobii TX300 & 300 & yes  & 22 \\

 CEM-\begin{CJK*}{UTF8}{gbsn}Ⅱ\end{CJK*}\cite{Holland2012BiometricVV} & EyeLink 1000 & 1000 & yes & 32 \\
 VREM\cite{Lohr2020EyeMB} & ET-HMD & 250 & yes  & 68 \\
 SBA-ST\cite{Friedman2016MethodTA} &EyeLink 1000 & 1000 & yes  & 298 \\ [1ex] 
\bottomrule
 \end{tabular}
 \label{table1}
\end{table}

\begin{figure}[htbp]
    \centerline{\includegraphics[scale=0.45]{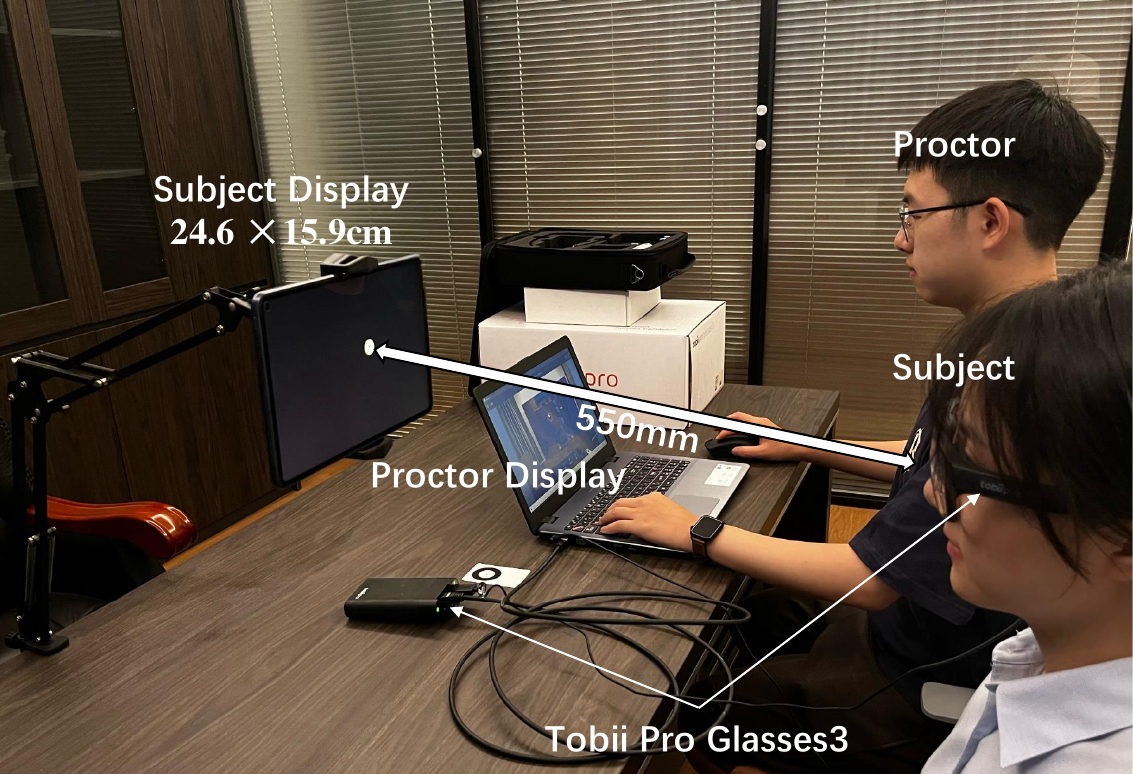}}
    \caption{Data collecting}
    \label{fig5}
\end{figure}

\paragraph{EMglasses}
{Existing eye movement datasets are collected at a sampling frequency of more than 250 Hz with the head posture fixed during the capturing processing (Table \ref{table1}). Such data with high-frequency and low-noise data can achieve good recognition result, it is hard to achieve recognition in practical application. For example, the is not  acceptable for user to fixed the head for long time and the data with high-frequency requires more storage and computation cost. In fact, the  frequency of commonly used camera  is usually between 30Hz and 60Hz and the work \cite{Holland2012BiometricVV} shows that the signal with sampling frequency of more than 30Hz can meet the requirements of eye movement recognition.} Therefore, we have built a challenging and realistic dataset to promote the development of eye movement recognition. Eye movements were captured at a 50 Hz sampling rate using a low cost device, Tobii Pro Glasses3 eye, a lightweight and unobtrusive tracker, allowing users to wear it comfortably for extended periods of time, and offering the advantage of mobility in real-world environments. We collected binocular eye movement data from 203 volunteers, with age ranging from 19 to 39 years old, where 85 $\%$ volunteers were myopic. Participants were seated in front of a 24.6 $\times$ 15.9 cm tablet, 50 - 60 cm from the display. The tablet was placed on a adjustable stand to ensure that the line of sight was perpendicular to the display at different heights (as shown in Fig. \ref{fig5}). Compared to  existing datasets, participants' heads were not stabilized using a chin and forehead rest (Table 3), and slight shaking was acceptable making, therefore, our dataset more representative for realistic real-life scenarios.

\begin{figure}[htbp]
    \centerline{\includegraphics[scale=0.45]{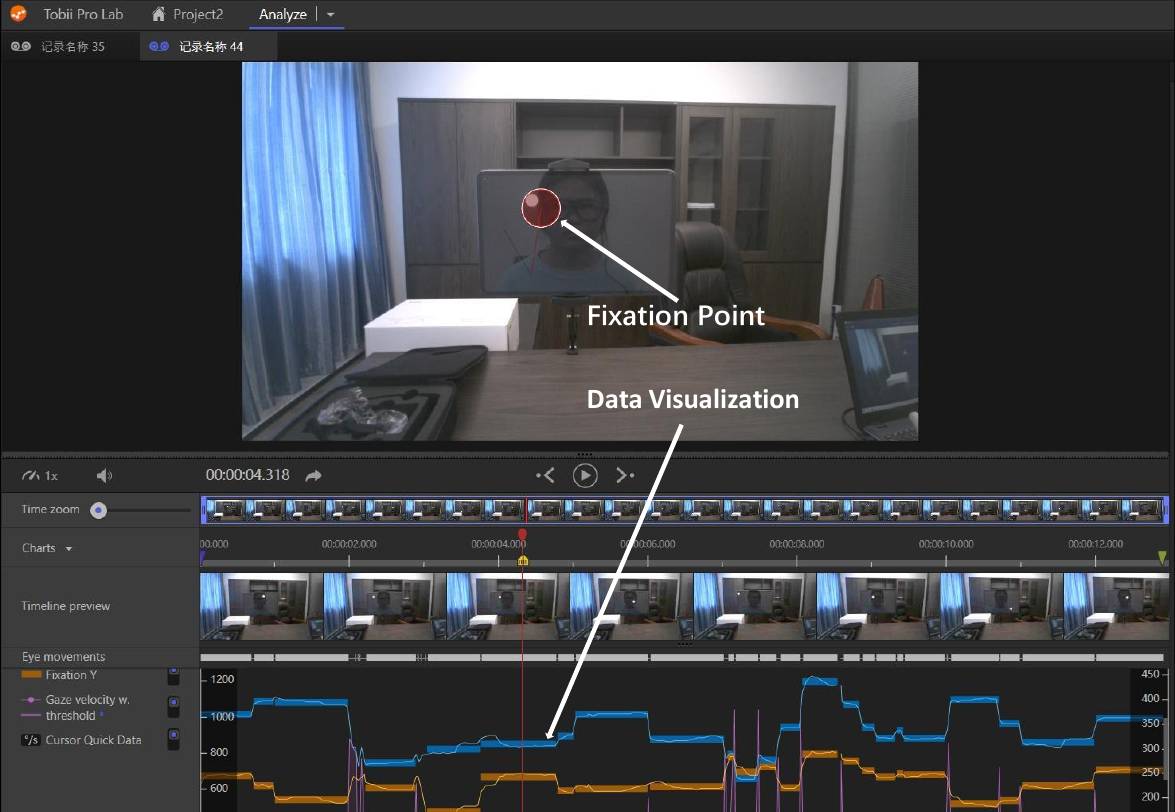}}
    \caption{Collected data.}
    \label{fig6}
\end{figure}

In each experiment, a black-white ball with a diameter of 1 cm, displayed as a stimulus, was randomly moved at the screen and stopped for 1 second in each position. Participants tracked the ball for about 10 seconds by moving their gaze point. During each recording, the gaze location was monitored by the proctor to ensure compliance. As shown in Fig.\ref{fig6}, the red circle represents the position of the volunteer's gaze point, while the curves below are the time-series visualization displays of eye movement. Participants completed data collection from two sessions, with the proctor suggesting to participants to take a few minutes break between each session if needed.  As a result, 406 (2 sessions $\times$ 203 participants) sequences from 203 participants were collected.

\paragraph{GazeBase}
The GazeBase dataset \cite{Griffith2020GazeBaseAL} comprises eye-movement recordings from 322 volunteers using the EyeLink1000 eye-tracker instrument. Volunteers' heads and chins were immobilized at a distance of 550 mm from a 47.4 $\times$ 29.7 cm screen during data acquisition. Each volunteer was exposed to seven tasks with different stimulus materials during acquisition, namely a randomized scanning task (RAN), a reading task (TEX), two video viewing tasks (VD1 and VD2), a gaze task (FXS), a horizontal scanning task (HSS), and an eye-driven game task (BLG). Nine recording rounds were conducted over 37 months, where the participants in each round were recruited exclusively from the previous round. In our experiments, four sub-datasets in GazeBase, RAN, TEX, FXS, and HSS, are used for recognition.
\paragraph{JuDo1000}
The JuDo1000 dataset \cite{Makowski2021DeepEyedentificationLiveOB} contains eye movement data from 150 volunteers, each participating in four experimental sessions with a time interval of more than one week between any two sessions. The data were recorded using an EyeLink Portable Duo eye tracker with a sampling frequency of 1,000 Hz.  The stimulus was a black dot with a diameter of 0.59 cm that appeared at five random locations on the screen during the experiment, with duration varying between 250, 500, and 1000 ms. Participants were seated in front of a 38×30 cm computer monitor at a height adjustable table and a viewing distance of 68 cm, with their head stabilized by a chin and forehead rest stand.

\subsection{Signal preprocessing}
As the scope of saccadic and fixational velocity are vastly different, global normalization would make the slow fixational drift and tremor to near zero, which results in missing key information in the eye tracking signal. Similar to works \cite{Makowski2021DeepEyedentificationLiveOB}\cite{Lohr2021EyeKY}, we split the original data into two subnets: a fast subnet and a slow subnet. Let X = ($x_{1},..., x_{t}$) and Y = ($y_{1},..., y_{t}$) be eye gaze point coordinate data. First, we transform the coordinate information into velocity information at each sampling point using following equations :
\begin{equation}
\delta^{(i)}_x = \frac{x^{(i)}-x^{(i-1)}}{t^{(i)}-t^{(i-1)}}
\end{equation}
\begin{equation}
\delta^{(i)}_y = \frac{y^{(i)}-y^{(i-1)}}{t^{(i)}-t^{(i-1)}}
\end{equation}

Then, we remove points without gaze data and replace the Nan values with 0. As a result, each eye data includes two channels: $\delta_{x}$ for the horizontal velocity sequence and $\delta_{y}$ for the vertical velocity sequence.

Furthermore, the preprocessing approach \cite{Makowski2021DeepEyedentificationLiveOB} \cite{Lohr2021EyeKY} is used to convert the raw data into two parts: fast data $D_{fast}$ and slow data $D_{slow}$. For the fast data, absolute velocities below a minimal
velocity $v_{min}$ are truncated and z-score normalization is employed for normalization.

\begin{equation}
D_{fast}^{(i)}{(\delta^i_x, \delta^i_y)}=\left\{\begin{matrix}
Z(\delta^i_x),Z(\delta^i_y)\quad\quad\quad otherwise\\

\quad Z(0) \quad\quad if\sqrt{(\delta^i_x)^{2}+(\delta^i_y)^{2}}< v_{min}
\end{matrix}\right.
\end{equation}

\begin{equation}
Z_{i}=(\delta^{i}-\mu)/\sigma
\end{equation}
where $\delta^{i}$ is the velocity at the $i$-th sampling point, and $\mu$ and $\sigma$ are the mean and variance of the sequence $\delta$.

For slow data $D_{slow}$, we use the hyperbolic tangent function to transform the input into a range of (-1, +1):
\begin{equation}
D_{slow}^{(i)}{(\delta^i_x, \delta^i_y)} = (tanh(c\delta^i_x),tanh(c\delta^i_y))
\end{equation}

The velocity threshold $v_{min}$ = 40 and the scaling factor c = 0.02 are two fixed hyperparameters.

After preprocessing, the fast data and slow data are input to Siamese CNN for feature embedding (as shown in Fig.\ref{fig2}). The latter's outputs are further input to the mix block for feature extraction and classification.

\subsection{Experimental settings}
To assess our approach, we split each dataset into training and testing sets. In our dataset, comprising 406 sequences from 203 participants, we select data collected at the first session for training and the remaining data for testing. For the GazeBase dataset, we consider the four sub-datasets, RAN, TEX, FXS, and HSS, with each providing 9 rounds (Rounds 1-9) of data  collected over  37-months, and two collections per round. As participants in a round were recruited only from the previous round, the first round collection includes data from all participants. After Round 6, fewer than 60 participants remain. The data from the first and second sessions in the first round are considered for training and testing, respectively. As a result, the training and testing sets both include 322 sequences for the first round. Accordingly, the four sub-datasets, RAN, TEX, FXS, and HSS, result in four training sets and testing sets at the first round. To evaluate the generalization ability of our approach w.r.t processed data at long time intervals, round 1-6 data from the RAN dataset (as listed in Table \ref{table8}) is also considered for comparable experiments. Specifically, we used round 1 data collected from the first session as the training set and round 1-6 data collected from the second session as the testing set. The training and test sets both include 322 sequences (322 participants $\times$ 2)  at  the first round.  At  the second round, we have 136 participants that had also provided the eye movement data at the first round. So, we select the data from these 136 participants at the first round for training and the data from the same 136 participants at the second round for testing. Accordingly, the training and the testing sets both include 136 sequences. In the same way, we obtain the training and testing sets for the remaining 3-6 rounds. Table \ref{table8} shows information regarding the RAN sub-dataset. \begin{table}[!htbp]
\caption{The RAN database collects information from the first round to the sixth round.}
\centering
\scalebox{0.78}{
\begin{tabular}{c c c c c c c}
\toprule
\textbf{Information} & \textbf{Round1} & \textbf{Round2} & \textbf{Round3}& \textbf{Round4}& \textbf{Round5}& \textbf{Round6}\\
\midrule
Time interval(month)& 0 & 1 & 2& 2& 9& 15\\
Number of participants & 322 & 136 & 105 & 101 & 78 & 59\\
\bottomrule
\end{tabular}
}
\label{table8}
\end{table}


For JuDo1000, 600 sequences were collected from 150 participants, each providing data for four times. The data from the first three times are selected for training while the last collection is selected as a test set. Accordingly, there are 450 sequences (150 participants $\times$ 3 times) in the training set and 150 sequences (150 participants $\times$ 1 times) in the testing set.

In our dataset, 203 participants provided eye movement data from two sessions with few minutes interval at one round. The data from the first session are used for training and the data from the second session for testing. Accordingly, the training and testing sets both comprise 203 sequences.

 To optimize the network parameters, we used the cross-entropy loss function in conjunction with the Adam optimizer. The cross-entropy loss was used to quantify the disparity between the model's output and the actual labels, enabling thereby the model to learn correct classifications. The Adam optimizer dynamically adjusts the network parameters to continuously enhance model's performance during training. The learning rate was fixed to 0.0002 and the batch size set to 64. All models were trained for 1000 rounds.

To assess our approach, the Equal Error Rate (EER) is computed. EER indicates the error rate when the  False Acceptance Rate (FAR) and False Rejection Rate (FRR) are equal. Lower EER values indicate better verification performance. We also report the FRR at different FARs $10^{-1}$, $10^{-2}$, and $10^{-3}$.


\subsection{Experimental results for short time interval}
\begin{figure*}[htbp]
    \centerline{\includegraphics[scale=0.8]{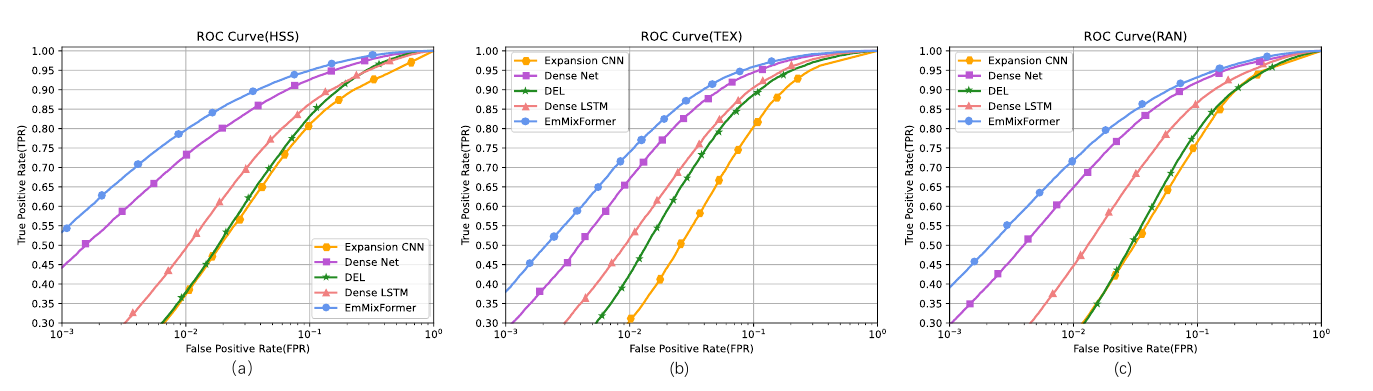}}
    \centerline{\includegraphics[scale=0.8]{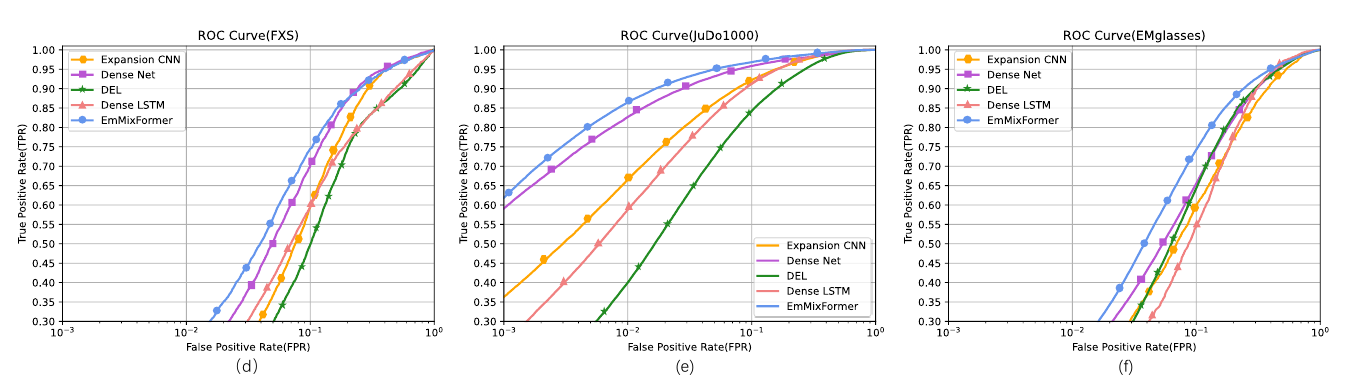}}
    \caption{ ROC Curve of various approaches on (a) HSS, (b) TEX, (c) RAN, (d) FXS in the GazeBase dataset,  (e) JuDo1000 dataset, and  (d) our dataset   }
    \label{fig7}
\end{figure*}

 These experiments aim to assess the verification performance of our approach on \textit{short} time interval data.  In the three datasets, the participants provided data from two sessions at each round. As detailed in section III-C, we select data from the first session for training and the remaining data for testing. The EMglasses dataset comprises 203 sequences in the training set and testing set, the four sub-datasets  (RAN, TEX, FXS, and HSS) in the the GazeBase dataset comprises 322 sequences in the training set and the testing set, while the JuDo1000 dataset comprises  450 sequences in the training set and 150 sequences in the testing set. Note that in all datasets, we split each sequence, collected for several seconds, into several sub-sequences for training and testing. The state of the art approaches, i.e. DEL\cite{Makowski2021DeepEyedentificationLiveOB}, Expansion CNN\cite{Lohr2021EyeKY}, Dense LSTM\cite{Taha2023EyeDriveAD} and DenseNet \cite{Lohr2022EyeKY}  are employed for benchmarking. The EER and FRR@FAR  of each approach are listed in Table \ref{table2},  Table \ref{table3}, Table \ref{table4}, Table \ref{table5}, Table \ref{table6}, and Table \ref{table7}.  The corresponding ROC curves are  shown in Fig. \ref{fig7}. It can be observed that our EmMixFormer outperforms the existing approaches and  achieves the lowest verification error, i.e. 0.0673, 0.0635, 0.0801, 0.1578 on the four subsets of the GazeBase dataset, 0.0502 on the JuDo1000 dataset, and  0.1599 on our dataset. Also,  we observe from Fig.\ref{fig7} that our proposed approach achieves the higher recognition accuracy w.r.t existing approaches at different FARs. 
\begin{table}[!htbp]
\caption{Results of comparative experiments on the HSS database}
\centering
\begin{tabular}{c c c c c}
\toprule
\multirow{2}*{HSS} &\multirow{2}*{EER} & \multicolumn{3}{c}{FRR@FAR}\\
\cline{3-5}
   &  & $10^{-1}$ & $10^{-2}$  & $10^{-3}$\\
\midrule
\textbf{EmMixformer} &\textbf{0.0673} &\textbf{0.0502} & \textbf{0.2032} & \textbf{0.4659} \\

 DEL\cite{Makowski2021DeepEyedentificationLiveOB} & 0.1309 & 0.1680  & 0.6217  & 0.9087 \\

Expansion CNN\cite{Lohr2021EyeKY} & 0.1437 & 0.1894 & 0.6267 & 0.9245 \\

Dense LSTM\cite{Taha2023EyeDriveAD} & 0.1191 & 0.1365 & 0.5082 & 0.8382\\

DenseNet\cite{Lohr2022EyeKY}& 0.0839 & 0.0739 & 0.2691  & 0.5575 \\
\bottomrule
\end{tabular}
\label{table2}
\end{table}
\begin{table}[!htbp]
\caption{Results of comparative experiments on the TEX database}
\centering
\begin{tabular}{c c c c c}
\toprule
\multirow{2}*{TEX} &\multirow{2}*{EER} & \multicolumn{3}{c}{FRR@FAR}\\
\cline{3-5}
   &  & $10^{-1}$ & $10^{-2}$  & $10^{-3}$\\
\midrule

\textbf{EmMixformer} & \textbf{0.0635} & \textbf{0.0407} & \textbf{0.2603} & \textbf{0.6193}\\
 DEL\cite{Makowski2021DeepEyedentificationLiveOB} & 0.1060 & 0.1128 & 0.5750 & 0.9141\\

Expansion CNN\cite{Lohr2021EyeKY} & 0.1362 & 0.1950 & 0.6977 & 1.0000\\

Dense LSTM\cite{Taha2023EyeDriveAD} & 0.0971 & 0.0945 & 0.4824 & 0.8456\\

DenseNet\cite{Lohr2022EyeKY}& 0.0736 & 0.0551 & 0.3293  & 0.7175 \\
\bottomrule
\end{tabular}
\label{table3}
\end{table}
\begin{table}[!htbp]
\caption{Results of comparative experiments on the RAN database}
\centering
\begin{tabular}{c c c c c}
\toprule
\multirow{2}*{RAN} &\multirow{2}*{EER} & \multicolumn{3}{c}{FRR@FAR}\\
\cline{3-5}
   &  & $10^{-1}$ & $10^{-2}$  & $10^{-3}$\\
\midrule
\textbf{EmMixformer} & \textbf{0.0801} & \textbf{0.0680} & \textbf{0.2818}  & \textbf{0.6087} \\

 DEL\cite{Makowski2021DeepEyedentificationLiveOB} & 0.1436 & 0.2066 & 0.7383  & 0.9645\\

Expansion CNN\cite{Lohr2021EyeKY} & 0.1500 & 0.2340 & 0.7277 & 0.9902\\

Dense LSTM\cite{Taha2023EyeDriveAD} & 0.1161 & 0.1329 & 0.5529 & 0.8846\\

DenseNet\cite{Lohr2022EyeKY}& 0.0885 & 0.0807 &  0.3513 & 0.7045 \\
\bottomrule
\end{tabular}
\label{table5}
\end{table}
\begin{table}[!htbp]
\caption{Results of comparative experiments on the FXS database}
\centering
\begin{tabular}{c c c c c}
\toprule
\multirow{2}*{FXS} &\multirow{2}*{EER} & \multicolumn{3}{c}{FRR@FAR}\\
\cline{3-5}
   &  & $10^{-1}$ & $10^{-2}$  & $10^{-3}$\\
\midrule
\textbf{EmMixformer} & \textbf{0.1578} & \textbf{0.2549} & \textbf{0.7654} & \textbf{0.9539} \\

 DEL\cite{Makowski2021DeepEyedentificationLiveOB} & 0.2229 & 0.5022  &  0.9358 & 0.9936\\

Expansion CNN\cite{Lohr2021EyeKY} & 0.1950 & 0.4051 & 0.9067 &0.9903 \\

Dense LSTM\cite{Taha2023EyeDriveAD} & 0.2191 & 0.4047 &0.8831  &0.9832 \\

DenseNet\cite{Lohr2022EyeKY}& 0.1666 & 0.2979 &0.8323  & 0.9821 \\
\bottomrule
\end{tabular}
\label{table6}
\end{table}
\begin{table}[!htbp]
\caption{Results of comparative experiments on the JuDo1000 database}
\centering
\begin{tabular}{c c c c c}
\toprule
\multirow{2}*{JuDo1000} &\multirow{2}*{EER} & \multicolumn{3}{c}{FRR@FAR}\\
\cline{3-5}
   &  & $10^{-1}$ & $10^{-2}$  & $10^{-3}$\\
\midrule
\textbf{EmMixformer}& 0.0543 & 0.0059 & 0.1284 & 0.3359\\

 DEL\cite{Makowski2021DeepEyedentificationLiveOB} & 0.1238 & 0.0781  &  0.5508 & 0.8945\\

Expansion CNN\cite{Lohr2021EyeKY} & 0.0989 & 0.0586 & 0.3203 & 0.7594\\

Dense LSTM\cite{Taha2023EyeDriveAD} & 0.0669 & 0.0195 & 0.2305 & 0.6016\\

DenseNet\cite{Lohr2022EyeKY}& 0.0773 & 0.0125 & 0.1953  & 0.4922 \\
\bottomrule
\end{tabular}
\label{table4}
\end{table}
\begin{table}[!htbp]
\caption{Results of comparative experiments on the EMglasses database}
\centering
\begin{tabular}{c c c c c}
\toprule
\multirow{2}*{EMglasses} &\multirow{2}*{EER} & \multicolumn{3}{c}{FRR@FAR}\\
\cline{3-5}
   &  & $10^{-1}$ & $10^{-2}$  & $10^{-3}$\\
\midrule
\textbf{EmMixformer}& \textbf{0.1599} & \textbf{0.2572}  & \textbf{0.7888} & \textbf{0.9711}\\

DEL\cite{Makowski2021DeepEyedentificationLiveOB} & 0.1853 & 0.3565  & 0.8877 & 0.9900\\

Expansion CNN\cite{Lohr2021EyeKY} & 0.2140 & 0.4013 & 0.8654 &0.9821 \\

Dense LSTM\cite{Taha2023EyeDriveAD} & 0.2069 & 0.4545 & 0.8928 &1.0000 \\

DenseNet\cite{Lohr2022EyeKY}& 0.1892 & 0.3444 & 0.8131  & 0.9812 \\
\bottomrule
\end{tabular}
\label{table7}
\end{table}

\subsection{Experimental results for long time intervals}
In the comparative experiments, we evaluated the authentication performance of our model on eye-tracking data collected at \textit{long} time intervals. The RAN sub-dataset in GazeBase comprises 1-6 round eye movement data over a 37-month collection cycle with the users providing data from two sessions for each round. As detailed in Section III-C, data collected at the first session in round 1 are used for model training, while data collected at the second session from round 1 to round 6 are used for test. Detailed information of Round 1-6 in the RAN dataset is given in Table II, where the time interval represents the duration in months of each round. The verification error rates are listed in Table \ref{table9}. We observe that our EmMixFormer consistently achieves the lowest EER for the different datasets collected at different times. It is worth noting also that all methods obtain higher verification error rates with increasing time intervals. \begin{table*}[!htbp]
\caption{Results of long-term interval experiments on the RAN database}
\centering
\scalebox{1.2}{  
\begin{tabular}{c c c c c c c}
\toprule
\multirow{2}*{Model} & \multicolumn{6}{c}{EER}\\
\cline{2-7}
 & Round1 & Round2 & Round3 & Round4 & Round5 & Round6 \\
\midrule
\textbf{EmMixformer} & \textbf{0.0801} & \textbf{0.1428} & \textbf{0.1803} & \textbf{0.1793} & \textbf{0.1853}  & \textbf{0.2279} \\

 DEL\cite{Makowski2021DeepEyedentificationLiveOB} & 0.1436 & 0.2215 & 0.2594 &0.2926 &0.2459 & 0.2907 \\

Expansion CNN\cite{Lohr2021EyeKY} & 0.1500 & 0.2328 & 0.2769 &0.2395& 0.2812 & 0.3057 \\

Dense LSTM\cite{Taha2023EyeDriveAD} & 0.1161 & 0.1939 & 0.2535 &0.2791 & 0.2322 & 0.2840 \\

DenseNet\cite{Lohr2022EyeKY} & 0.0885 & 0.1775 & 0.2161 & 0.2032 & 0.2047 & 0.2592 \\
\bottomrule
\end{tabular}
}
\label{table9}
\end{table*}

Overall, all the experimental results show that the proposed EmMixFormer significantly outperforms existing eye movement recognition approaches and achieves the lowest verification errors on the three datasets. Such good performance is explained by the following facts. The CNN based approaches, i.e. DEL\cite{Makowski2021DeepEyedentificationLiveOB}, Expansion CNN\cite{Lohr2021EyeKY}, and DenseNet\cite{Lohr2022EyeKY}, leverage the local receptive fields and shared weights to learn local structure, but are unable to learn robust features that embed long-range temporal dependencies. Dense LSTM\cite{Taha2023EyeDriveAD}, by contrast, is capable of capturing temporal dependencies in sequential data as memory cells can store information over a long period, allowing them to remember important information from earlier time steps. This capacity, nonetheless, is only partial owing to the shallow learning implemented by LSTM. Transformers, thanks to their self-attention mechanism for capturing dependencies between all positions in the input sequence, are much more effective in modeling long-term dependencies without the need for recurrent connections. On the other side, the self attention in frequency domain learns global dependencies for features representation.  By leveraging conjointly long-range time-dependencies and global dependencies, our EmMixFormer is much more powerful to model the eye movement data w.r.t other approaches, which explains the new state-of-the-art results it achieves on the data collected at long or short intervals. Besides, Table \ref{table9} shows that the performance of all approaches is degraded with increasing time duration. This is attributed to the fact that there are more variations in collected data with long time intervals, which results in increasing mismatching during recognition.

\subsection{Ablation experiment}
\begin{figure}[htbp]
    \centerline{\includegraphics[scale=1]{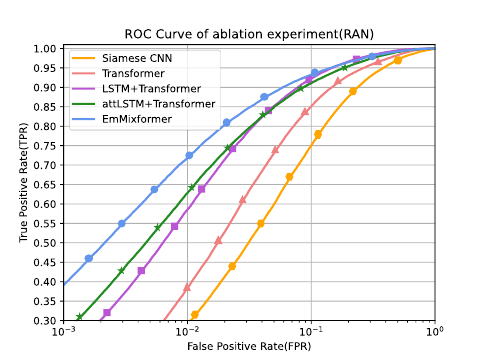}}
    \caption{ROC Curve of ablation experiment on RAN in the GazeBase dataset}
    \label{fig8}
\end{figure}

 Our our EmMixFormer consists of siamese CNN and Mix block, where Mix block includes three modules: Attention LSTM, Transformer, and Fourier Transformer (as shown in Fig.\ref{fig2}). To investigate the effect of each module w.r.t  recognition accuracy improvement, we conducted ablation experiments on the round 1 data from the RAN sub-dataset in GazeBase. To ease the description, we remove Mix block from EmMixFormer  to obtain the basic siamese CNN. Then, we gradually add  attention LSTM, Transformer, and Fourier Transformer, with the resulting models denoted as Siamese CNN+Transformer, Siamese CNN+LSTM+Transformer and Siamese CNN+attLSTM+Transformer, respectively.
 Afterwards, the Fourier Transformer, is included to the attLSTM+Transformer, with the resulting model represented as Siamese CNN+attLSTM+Transformer+Fourierformer, i.e. EmMixFormer. The verification error rates of these ablation schemes are shown in Table \ref{table10} and Fig.\ref{fig8}. The experimental results demonstrate that incorporating Transformer into siamese CNN increases the accuracy, which implies that Transformer can learn dependency features to improve recognition performance. Also, we can observe that the combination of the Attention LSTM and the Fourier Transformer allows to improve the
identification accuracy of simple Transformer. Besides, comparing Transformer with LSTM+Transformer, we observe that the LSTM can provide compensation features to improve the performance of Transformer, which is supported by the work \cite{huang2020trans}. From Table \ref{table10} and Fig.\ref{fig8}, we also observe that attLSTM+Transformer achieves lower EER than LSTM+Transformer, which implies that the attention mechanism improves LSTM performance. Finally, the performance becomes optimal when attLSTM+Transformer is combined with Fourier Transformer, which demonstrates that the attention mechanism in the frequency domain captures global dependencies for feature representation. 

\begin{table}[!htbp]
\caption{Results of the ablation experiments on the RAN database}
\centering
\begin{tabular}{c c c c c}
\toprule
\multirow{2}*{Method} &\multirow{2}*{EER} & \multicolumn{3}{c}{FRR@FAR}\\
\cline{3-5}
   &  & $10^{-1}$ & $10^{-2}$  & $10^{-3}$\\
\midrule
Siamese CNN & 0.1594 & 0.2498  &0.7091 & 0.9900\\

+Transformer & 0.1214 & 0.1491 & 0.6167 &0.9133 \\

LSTM+Transformer& 0.0880 & 0.0776 & 0.4148  & 0.8204 \\

attLSTM+Transformer & 0.0936 & 0.0887 & 0.3720 &0.7377 \\

\textbf{EmMixformer}& \textbf{0.0801} & \textbf{0.0680}  & \textbf{0.2818} & \textbf{0.6087}\\
\bottomrule
\end{tabular}
\label{table10}
\end{table}

\section{Conclusion}
This paper proposes an end-to-end model, termed EmMixFormer, for eye movement biometric authentication. EmMixFormer combines Attention LSTM, Transformer, and Fourier transformer for feature representation. As the attention LSTM and Transformer are  capable of learning temporal dependencies within eye movement data while Fourier transformer can capture global feature dependencies, our proposed EmMixFormer is able to infer feature representation embeddings that combine long-range time-dependencies and global dependencies. Our experimental results on three representative datasets show that our approach outperforms the existing approaches and achieves a new state-of-the-art verification accuracy.

In the future, we will focus on the following research directions. First, we will collect more data to build larger datasets for evaluating eye-tracking models. This will provide a broader and more diverse dataset to improve performance. Second, we will refine the preprocessing methods and model architectures for eye-tracking data to enhance feature representation. Additionally, since eye movements are independent of existing biometric modalities, we can explore integrating eye movements as a supplementary biometric feature with other modalities. This fusion scheme can be leveraged in multimodal biometric systems to enhance system recognition accuracy and security.


\bibliographystyle{unsrt}
\bibliography{TEX}

\begin{thebibliography}{10}

\bibitem{Jain2008AnIT}
Anil~K. Jain, Arun Ross, and Karthik Nandakumar.
\newblock An introduction to biometrics.
\newblock In {\em International Conference on Pattern Recognition}, 2008.

\bibitem{Qin2023LabelEM}
Huafeng Qin, Changqing Gong, Yantao Li, Xinbo Gao, and Moun{\^i}m~A. El-Yacoubi.
\newblock Label enhancement-based multiscale transformer for palm-vein recognition.
\newblock {\em IEEE Transactions on Instrumentation and Measurement}, 72:1--17, 2023.

\bibitem{Jirachaweng2011ResidualOM}
Suksan Jirachaweng, Zujun Hou, Wei-Yun Yau, and Vutipong Areekul.
\newblock Residual orientation modeling for fingerprint enhancement and singular point detection.
\newblock {\em Pattern Recognit.}, 44:431--442, 2011.

\bibitem{Wang2010FaceRU}
Yong Wang and Yi~Wu.
\newblock Face recognition using intrinsicfaces.
\newblock {\em Pattern Recognit.}, 43:3580--3590, 2010.

\bibitem{Qin2011RegionGF}
Huafeng Qin, Lan Qin, and Chengbo Yu.
\newblock Region growth-based feature extraction method for finger-vein recognition.
\newblock {\em Optical Engineering}, 50:057208, 2011.

\bibitem{Roy2011TowardsNI}
Kaushik Roy, Prabir Bhattacharya, and Ching~Yee Suen.
\newblock Towards nonideal iris recognition based on level set method, genetic algorithms and adaptive asymmetrical svms.
\newblock {\em Eng. Appl. Artif. Intell.}, 24:458--475, 2011.

\bibitem{Lohr2021EyeKY}
Dillon~James Lohr, Henry~K. Griffith, and Oleg~V. Komogortsev.
\newblock Eye know you: Metric learning for end-to-end biometric authentication using eye movements from a longitudinal dataset.
\newblock {\em IEEE Transactions on Biometrics, Behavior, and Identity Science}, 4:276--288, 2021.

\bibitem{Cola2016GaitbasedAU}
Guglielmo Cola, Marco Avvenuti, Fabio Musso, and Alessio Vecchio.
\newblock Gait-based authentication using a wrist-worn device.
\newblock {\em Proceedings of the 13th International Conference on Mobile and Ubiquitous Systems: Computing, Networking and Services}, 2016.

\bibitem{Kiran2021OfflineSR}
P.~Kiran, Bidare Parameshachari, J.~Yashwanth, and K.~Bharath.
\newblock Offline signature recognition using image processing techniques and back propagation neuron network system.
\newblock {\em SN Computer Science}, 2, 2021.

\bibitem{Liang2020BehavioralBF}
Yunji Liang, Sagar Samtani, Bin Guo, and Zhiwen Yu.
\newblock Behavioral biometrics for continuous authentication in the internet-of-things era: An artificial intelligence perspective.
\newblock {\em IEEE Internet of Things Journal}, 7:9128--9143, 2020.

\bibitem{Chugh2018FingerprintSB}
T.~Chugh, Kai Cao, and Anil~K. Jain.
\newblock Fingerprint spoof buster: Use of minutiae-centered patches.
\newblock {\em IEEE Transactions on Information Forensics and Security}, 13:2190--2202, 2018.

\bibitem{Makowski2020BiometricIA}
Silvia Makowski, Lena~A. J{\"a}ger, Paul Prasse, and Tobias Scheffer.
\newblock Biometric identification and presentation-attack detection using micro- and macro-movements of the eyes.
\newblock {\em 2020 IEEE International Joint Conference on Biometrics (IJCB)}, pages 1--10, 2020.

\bibitem{Holland2013ComplexEM}
Corey Holland and Oleg~V. Komogortsev.
\newblock Complex eye movement pattern biometrics: Analyzing fixations and saccades.
\newblock {\em 2013 International Conference on Biometrics (ICB)}, pages 1--8, 2013.

\bibitem{Blignaut2013MappingTP}
Pieter~J. Blignaut.
\newblock Mapping the pupil-glint vector to gaze coordinates in a simple video-based eye tracker.
\newblock {\em Journal of Eye Movement Research}, 7, 2013.

\bibitem{Gunetti2005KeystrokeAO}
Daniele Gunetti and Claudia Picardi.
\newblock Keystroke analysis of free text.
\newblock {\em ACM Trans. Inf. Syst. Secur.}, 8:312--347, 2005.

\bibitem{Zheng2011AnEU}
Nan Zheng, Aaron Paloski, and Haining Wang.
\newblock An efficient user verification system via mouse movements.
\newblock In {\em Conference on Computer and Communications Security}, 2011.

\bibitem{George2016ASL}
Anjith George and Aurobinda Routray.
\newblock A score level fusion method for eye movement biometrics.
\newblock {\em Pattern Recognit. Lett.}, 82:207--215, 2016.

\bibitem{noton1971scanpaths}
David Noton and Lawrence Stark.
\newblock Scanpaths in eye movements during pattern perception.
\newblock {\em Science}, 171(3968):308--311, 1971.

\bibitem{Nagamatsu2008OnepointCG}
Takashi Nagamatsu, Junzo Kamahara, Takumi Iko, and Naoki Tanaka.
\newblock One-point calibration gaze tracking based on eyeball kinematics using stereo cameras.
\newblock {\em Proceedings of the 2008 symposium on Eye tracking research \& applications}, 2008.

\bibitem{Akkil2014TraQuMeAT}
Deepak Akkil, Poika Isokoski, Jari Kangas, Jussi Rantala, and R.~Raisamo.
\newblock Traqume: a tool for measuring the gaze tracking quality.
\newblock {\em Proceedings of the Symposium on Eye Tracking Research and Applications}, 2014.

\bibitem{Kasprowski2004EyeMI}
Paweł Kasprowski and J{\'o}zef Ober.
\newblock Eye movements in biometrics.
\newblock In {\em ECCV Workshop BioAW}, 2004.

\bibitem{bednarik2005eye}
Roman Bednarik, Tomi Kinnunen, Andrei Mihaila, and Pasi Fr{\"a}nti.
\newblock Eye-movements as a biometric.
\newblock In {\em Image Analysis: 14th Scandinavian Conference, SCIA 2005, Joensuu, Finland, June 19-22, 2005. Proceedings 14}, pages 780--789. Springer, 2005.

\bibitem{Holland2012BiometricVV}
Corey Holland and Oleg~V. Komogortsev.
\newblock Biometric verification via complex eye movements: The effects of environment and stimulus.
\newblock {\em 2012 IEEE Fifth International Conference on Biometrics: Theory, Applications and Systems (BTAS)}, pages 39--46, 2012.

\bibitem{Rigas2018StudyOA}
Ioannis Rigas, Lee Friedman, and Oleg~V. Komogortsev.
\newblock Study of an extensive set of eye movement features: Extraction methods and statistical analysis.
\newblock {\em Journal of Eye Movement Research}, 11, 2018.

\bibitem{Holland2011BiometricIV}
Corey Holland and Oleg~V. Komogortsev.
\newblock Biometric identification via eye movement scanpaths in reading.
\newblock {\em 2011 International Joint Conference on Biometrics (IJCB)}, pages 1--8, 2011.

\bibitem{Komogortsev2012BiometricAV}
Oleg~V. Komogortsev, Alexey Karpov, Larry~R. Price, and Cecilia~R. Aragon.
\newblock Biometric authentication via oculomotor plant characteristics.
\newblock {\em 2012 5th IAPR International Conference on Biometrics (ICB)}, pages 413--420, 2012.

\bibitem{Bayat2017BiometricIT}
Akram Bayat and Marc Pomplun.
\newblock Biometric identification through eye-movement patterns.
\newblock In {\em International Conference on Applied Human Factors and Ergonomics}, 2017.

\bibitem{Nguyen2012MelfrequencyCC}
Cuong~V Nguyen, Vu~C. Dinh, and Lam Si~Tung Ho.
\newblock Mel-frequency cepstral coefficients for eye movement identification.
\newblock {\em 2012 IEEE 24th International Conference on Tools with Artificial Intelligence}, 1:253--260, 2012.

\bibitem{li2018biometric}
Chunyong Li, Jiguo Xue, Cheng Quan, Jingwei Yue, and Chenggang Zhang.
\newblock Biometric recognition via texture features of eye movement trajectories in a visual searching task.
\newblock {\em PloS one}, 13(4):e0194475, 2018.

\bibitem{lohr2020metric}
Dillon Lohr, Henry Griffith, Samantha Aziz, and Oleg Komogortsev.
\newblock A metric learning approach to eye movement biometrics.
\newblock In {\em 2020 IEEE International Joint Conference on Biometrics (IJCB)}, pages 1--7. IEEE, 2020.

\bibitem{yeamkuan2020fixational}
Suparat Yeamkuan and Kosin Chamnongthai.
\newblock Fixational feature-based gaze pattern recognition using long short-term memory.
\newblock In {\em 2020 Asia-Pacific Signal and Information Processing Association Annual Summit and Conference (APSIPA ASC)}, pages 1--4. IEEE, 2020.

\bibitem{Jia2018BiometricRT}
Shaohua Jia, Do~Hyong Koh, Amanda Seccia, Pasha Antonenko, Richard~L. Lamb, Andreas Keil, Matthew~H. Schneps, and Marc Pomplun.
\newblock Biometric recognition through eye movements using a recurrent neural network.
\newblock {\em 2018 IEEE International Conference on Big Knowledge (ICBK)}, pages 57--64, 2018.

\bibitem{Shelhamer2014FullyCN}
Evan Shelhamer, Jonathan Long, and Trevor Darrell.
\newblock Fully convolutional networks for semantic segmentation.
\newblock {\em 2015 IEEE Conference on Computer Vision and Pattern Recognition (CVPR)}, pages 3431--3440, 2014.

\bibitem{Krizhevsky2012ImageNetCW}
Alex Krizhevsky, Ilya Sutskever, and Geoffrey~E. Hinton.
\newblock Imagenet classification with deep convolutional neural networks.
\newblock {\em Communications of the ACM}, 60:84 -- 90, 2012.

\bibitem{Girshick2013RichFH}
Ross~B. Girshick, Jeff Donahue, Trevor Darrell, and Jitendra Malik.
\newblock Rich feature hierarchies for accurate object detection and semantic segmentation.
\newblock {\em 2014 IEEE Conference on Computer Vision and Pattern Recognition}, pages 580--587, 2013.

\bibitem{Taha2023EyeDriveAD}
Bilal Taha, Sherif Nagib~Abbas Seha, Dae~Yon Hwang, and Dimitrios Hatzinakos.
\newblock Eyedrive: A deep learning model for continuous driver authentication.
\newblock {\em IEEE Journal of Selected Topics in Signal Processing}, 17:637--647, 2023.

\bibitem{Jger2019DeepEB}
Lena~A. J{\"a}ger, Silvia Makowski, Paul Prasse, Sascha Liehr, Maximilian Seidler, and Tobias Scheffer.
\newblock Deep eyedentification: Biometric identification using micro-movements of the eye.
\newblock {\em ArXiv}, abs/1906.11889, 2019.

\bibitem{Makowski2021DeepEyedentificationLiveOB}
Silvia Makowski, Paul Prasse, David~Robert Reich, Daniel~G. Krakowczyk, Lena~A. J{\"a}ger, and Tobias Scheffer.
\newblock Deepeyedentificationlive: Oculomotoric biometric identification and presentation-attack detection using deep neural networks.
\newblock {\em IEEE Transactions on Biometrics, Behavior, and Identity Science}, 3:506--518, 2021.

\bibitem{Makowski2022OculomotoricBI}
Silvia Makowski, Paul Prasse, Lena~A. J{\"a}ger, and Tobias Scheffer.
\newblock Oculomotoric biometric identification under the influence of alcohol and fatigue.
\newblock {\em 2022 IEEE International Joint Conference on Biometrics (IJCB)}, pages 1--9, 2022.

\bibitem{Lohr2022EyeKY}
Dillon~James Lohr and Oleg~V. Komogortsev.
\newblock Eye know you too: A densenet architecture for end-to-end biometric authentication via eye movements.
\newblock {\em ArXiv}, abs/2201.02110, 2022.

\bibitem{Guo_2022_CVPR}
Jianyuan Guo, Kai Han, Han Wu, Yehui Tang, Xinghao Chen, Yunhe Wang, and Chang Xu.
\newblock Cmt: Convolutional neural networks meet vision transformers.
\newblock In {\em Proceedings of the IEEE/CVF Conference on Computer Vision and Pattern Recognition (CVPR)}, pages 12175--12185, June 2022.

\bibitem{Vaswani2017AttentionIA}
Ashish Vaswani, Noam~M. Shazeer, Niki Parmar, Jakob Uszkoreit, Llion Jones, Aidan~N. Gomez, Lukasz Kaiser, and Illia Polosukhin.
\newblock Attention is all you need.
\newblock In {\em NIPS}, 2017.

\bibitem{Dosovitskiy2020AnII}
Alexey Dosovitskiy, Lucas Beyer, Alexander Kolesnikov, Dirk Weissenborn, Xiaohua Zhai, Thomas Unterthiner, Mostafa Dehghani, Matthias Minderer, Georg Heigold, Sylvain Gelly, Jakob Uszkoreit, and Neil Houlsby.
\newblock An image is worth 16x16 words: Transformers for image recognition at scale.
\newblock {\em ArXiv}, abs/2010.11929, 2020.

\bibitem{hochreiter2001gradient}
Sepp Hochreiter, Yoshua Bengio, Paolo Frasconi, J{\"u}rgen Schmidhuber, et~al.
\newblock Gradient flow in recurrent nets: the difficulty of learning long-term dependencies, 2001.

\bibitem{Liu2021SwinTH}
Ze~Liu, Yutong Lin, Yue Cao, Han Hu, Yixuan Wei, Zheng Zhang, Stephen Lin, and Baining Guo.
\newblock Swin transformer: Hierarchical vision transformer using shifted windows.
\newblock {\em 2021 IEEE/CVF International Conference on Computer Vision (ICCV)}, pages 9992--10002, 2021.

\bibitem{Hochreiter1997LongSM}
Sepp Hochreiter and J{\"u}rgen Schmidhuber.
\newblock Long short-term memory.
\newblock {\em Neural Computation}, 9:1735--1780, 1997.

\bibitem{Ordonez2016DeepCA}
Francisco~Javier Ordonez and Daniel Roggen.
\newblock Deep convolutional and lstm recurrent neural networks for multimodal wearable activity recognition.
\newblock {\em Sensors (Basel, Switzerland)}, 16, 2016.

\bibitem{Ngo2017SaccadeGP}
Thuyen Ngo and B.~S. Manjunath.
\newblock Saccade gaze prediction using a recurrent neural network.
\newblock {\em 2017 IEEE International Conference on Image Processing (ICIP)}, pages 3435--3439, 2017.

\bibitem{Greff2015LSTMAS}
Klaus Greff, Rupesh~Kumar Srivastava, Jan Koutn{\'i}k, Bas~R. Steunebrink, and J{\"u}rgen Schmidhuber.
\newblock Lstm: A search space odyssey.
\newblock {\em IEEE Transactions on Neural Networks and Learning Systems}, 28:2222--2232, 2015.

\bibitem{Gers2000LearningTF}
Felix~Alexander Gers, J{\"u}rgen Schmidhuber, and Fred Cummins.
\newblock Learning to forget: Continual prediction with lstm.
\newblock {\em Neural Computation}, 12:2451--2471, 2000.

\bibitem{Li2023LocalGlobalTE}
Mingsong Li, Yikun Liu, Tao Xiao, Yuwen Huang, and Gong-Ping Yang.
\newblock Local-global transformer enhanced unfolding network for pan-sharpening.
\newblock {\em ArXiv}, abs/2304.14612, 2023.

\bibitem{Zhou2022DeepFU}
Man Zhou, Huikang Yu, Jie Huang, Fengmei Zhao, Jinwei Gu, Chen~Change Loy, Deyu Meng, and Chongyi Li.
\newblock Deep fourier up-sampling.
\newblock {\em ArXiv}, abs/2210.05171, 2022.

\bibitem{Griffith2020GazeBaseAL}
Henry~K. Griffith, Dillon~James Lohr, Evgeniy Abdulin, and Oleg~V. Komogortsev.
\newblock Gazebase, a large-scale, multi-stimulus, longitudinal eye movement dataset.
\newblock {\em Scientific Data}, 8, 2020.

\bibitem{Lohr2020EyeMB}
Dillon~James Lohr, Samantha Aziz, and Oleg~V. Komogortsev.
\newblock Eye movement biometrics using a new dataset collected in virtual reality.
\newblock {\em ACM Symposium on Eye Tracking Research and Applications}, 2020.

\bibitem{Friedman2016MethodTA}
Lee Friedman, Mark~S. Nixon, and Oleg~V. Komogortsev.
\newblock Method to assess the temporal persistence of potential biometric features: Application to oculomotor, gait, face and brain structure databases.
\newblock {\em PLoS ONE}, 12, 2016.

\bibitem{huang2020trans}
Zhiheng Huang, Peng Xu, Davis Liang, Ajay Mishra, and Bing Xiang.
\newblock Trans-blstm: Transformer with bidirectional lstm for language understanding.
\newblock {\em arXiv preprint arXiv:2003.07000}, 2020.

\end{thebibliography}

\end{document}